\documentclass[letterpaper, 10 pt, conference]{./ieee/ieeeconf}  

\IEEEoverridecommandlockouts                              

\usepackage{xcolor}
\usepackage{xspace}
\usepackage{booktabs}
\usepackage{colortbl}
\usepackage{multirow}
\usepackage{makecell}
\usepackage{lipsum} 
\usepackage{gensymb} 
\usepackage{textcomp, gensymb}  

\newcommand{\Tref}[1]{Table~\ref{#1}}
\newcommand{\eref}[1]{Eq.~\eqref{#1}}

\newcommand{\fref}[1]{Fig.~\ref{#1}}
\newcommand{\Fref}[1]{Figure~\ref{#1}}

\renewcommand{\paragraph}[1]{\vspace{0.2em}\noindent \textbf{#1 \hspace{0.2em}}}

\definecolor{MyDarkRed}{rgb}{0.46, 0.16, 0.16}
\definecolor{MyDarkBlue}{rgb}{0.16, 0.16, 0.66}

\usepackage{url}
\usepackage[backref]{hyperref}
\hypersetup{hidelinks}
\usepackage{xcolor}
\usepackage{graphicx}
\usepackage{amsmath} 
\usepackage{amssymb}  

\definecolor{darkblue}{RGB}{50,65,155}

\title{SymmeTac: 
Symmetric Color LED Driven Efficient Photometric Stereo \\
Reconstruction Methods for Camera-based Tactile Sensors
}

\author{Jieji Ren$^{1*}$, Heng Guo$^{2*}$, Zaiyan Yang$^{2}$, Jinnuo Zhang$^{1}$, Yueshi Dong$^{1}$, Ningbin Zhang$^{1}$, \\
Boxin Shi$^{3}$, Jiang Zou$^{1}$, Guoying Gu$^{1**}$
\thanks{This work was supported in part by NSFC No.52025057, 52275024.}
\thanks{$^{*}$Equally contributed; $^{**}$Corresponding author: Guoying Gu.}
\thanks{$^{1}$Jieji Ren, Jinnuo Zhang, Yueshi Dong, Ningbin Zhang, Jiang Zou and Guoying Gu are with the School of Mechanical Engineering, Shanghai Jiao Tong University, Shanghai, 200240, China. Email: guguoying@sjtu.edu.cn}%
\thanks{$^{2}$Heng Guo and Zaiyan Yang are with the School of Artificial Intelligence, Beijing University of Posts and Telecom., Beijing, 100876, China.}
\thanks{$^{3}$Boxin Shi is with the School of Computer Science, Peking University, Beijing, 100871, China.}
}%

\begin{document}

\maketitle
\thispagestyle{empty}
\pagestyle{empty}

\begin{abstract}
Camera-based tactile sensors can provide high-density surface geometry and force information for robots in the interaction process with the target. 
However, most existing methods cannot achieve accurate reconstruction with high efficiency, impeding the applications in robots.
To address these problems, we propose an efficient two-shot photometric stereo method based on symmetric color LED distribution. 
Specifically, based on the sensing response curve of CMOS channels, we design orthogonal red and blue LEDs as illumination to acquire four observation maps using channel-splitting in a two-shot manner.
Subsequently, we develop a two-shot photometric stereo theory, which can estimate accurate surface normal and greatly reduce the computing overhead in magnitude. 
Finally, leveraging the characteristics of the camera-based tactile sensor, we optimize the algorithm to be a highly efficient, pure addition operation.
Simulation and real-world experiments demonstrate the advantages of our approach. 
Further details are available on: \url{https://github.com/Tacxels/SymmeTac}.
\end{abstract}



\section{INTRODUCTION}
Tactile perception~\cite{liu2022neuro} is significant and indispensable for robots, especially for dexterous grasping and delicate manipulation in industry and daily life. 
Unlike classical electric-based tactile sensors~\cite{liu2022neuro}, camera-based tactile sensor overcomes the challenge of density, signal stability and processing~\cite{rodriguez2021unstable,bauza2024SimPLE}, and has attracted widespread attentions. 

Camera-based tactile sensors utilize camera to observe the contact deformation of its soft layer under illuminations, reconstruct the geometry of target surface and further estimate the contact force distribution, which can provide high-density, stable and multi-modality information with low-cost hardware~\cite{li2024tactileReveiw}.
With the help of advanced algorithms, it can support rich downstream robot tasks~\cite{bauza2024SimPLE,zhao2016optoelectronically,andrussow2023minsight} and drive the development of embodied AI~\cite{lepora2024future}. 
Fast and accurate geometric reconstruction algorithm is significant for camera-based tactile sensors~\cite{fang2024skin}, which not only provides shape and texture of target surface~\cite{yuan2017gelsight}, but also builds the basis for subsequent force estimation with mechanical model~\cite{zhao2023situ}. 
However, efficiently acquiring accurate surface geometries is still facing challenges for camera-based tactile sensors. 

There are mainly two approaches for camera-based tactile sensors to calculate the surface geometries. 
Model-based methods take photometric stereo (PS) and variants to accurately reconstruct the deformation surface of sensors~\cite{johnson2009retrographicOriPS}. However, the sequential observation process under different monochromatic lights seriously decreases the efficiency. 
Look-up table methods take chromatic lights (R,G,B) to illuminate the deformation surface from different directions, and build the mapping relationship between the observed intensity and surface gradient $(G_{ix}, G_{iy}) = \mathcal{F}(r_i,g_i,b_i)$ for each pixel $i$~\cite{li2014localizationGelsight,dong2017improvedGelsight}. 
Nevertheless, the table calibration is easily disturbed by imaging error or edge detection errors~\cite{yuan2017gelsight}, and the nonuniform illumination strongly affects the accuracy. If one wants to increase the accuracy, the table size will greatly expand and decrease the reconstruction accuracy. How to simultaneously achieve accurate, general and efficient reconstruction is still one of the biggest challenges of camera-based tactile sensors.

\begin{figure}[!t]
    \setlength{\abovecaptionskip}{-0.05cm} 
    \centering
    \includegraphics[height=1.02\columnwidth]{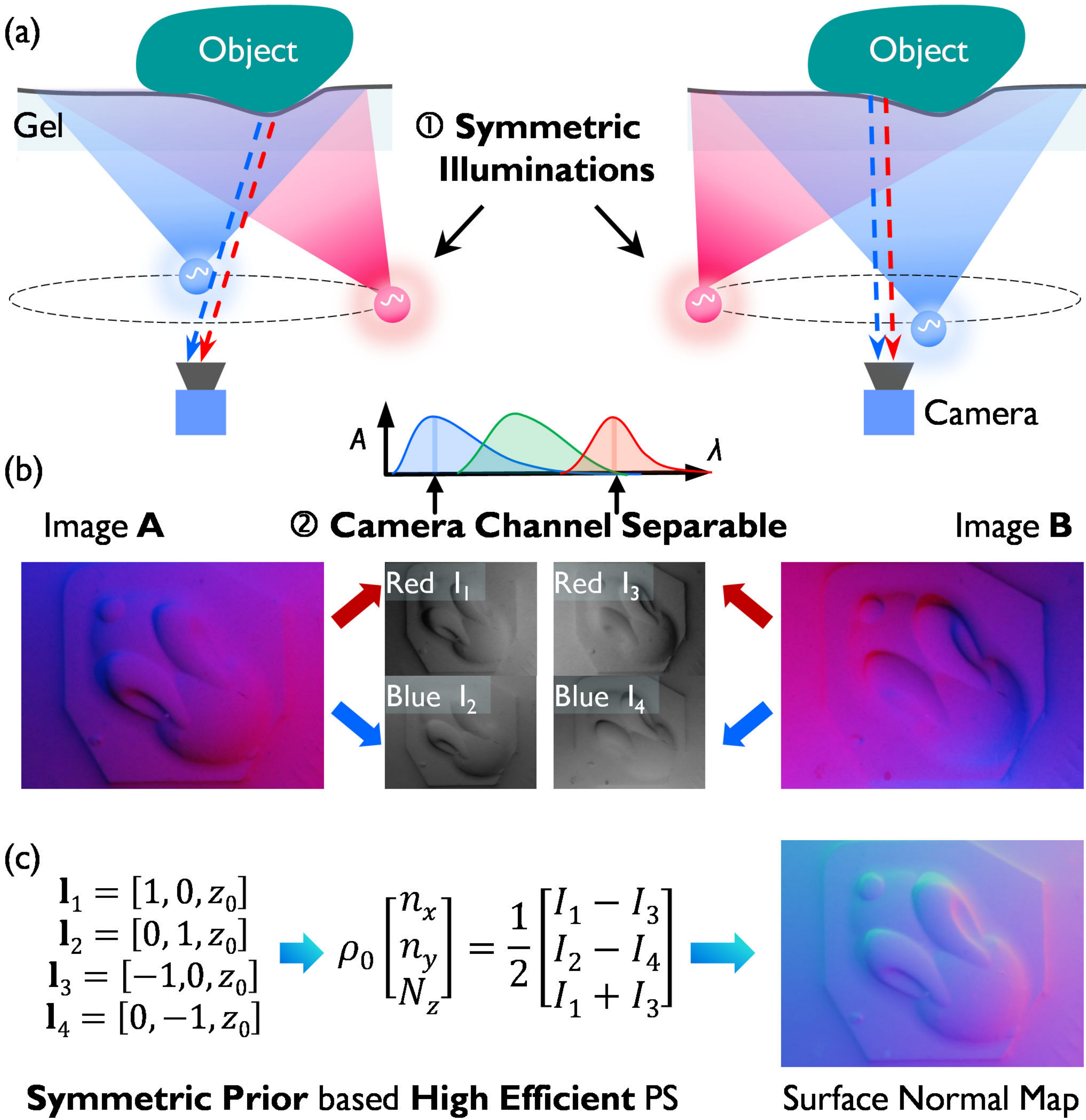}
    \caption{(a) Camera-based tactile sensor with symmetric illuminations. (b) Camera response curve supports channel splitting. (c) Proposed two-shot, highly efficient PS method and the reconstructed surface normal map.\vspace{-0.6cm}}
    \label{fig:teaser}
\end{figure}

To address these problems, we propose SymmeTac, a highly efficient PS reconstruction methods for camera-based tactile sensors with colorful symmetric illuminations, as shown in Fig. \ref{fig:teaser}.
Inspired by the response feature of the color camera CMOS, we could effectively split the Red channel and Blue channel as two independent observations under red and blue LEDs separately.
Then, we design the illumination as two group orthogonal and symmetric colorful lights, each group has a Red and a Blue LED separated by 90 degrees, which can provide 4 circumferential observations with only 2 lighting.
Finally, we build a two-shot photometric stereo model to reconstruct the surface normal, which can model the albedo of each wavelength band and achieve highly efficient reconstruction with simple calculations.
With proposed illumination design and model, camera-based tactile sensors can reconstruct surface with general and accurate model-based method, and more importantly, it is highly efficient.

The contributions of this work are concluded as follows:
\begin{itemize}
    \item A highly efficient two-shot PS method is proposed;
    \item The spectral characteristics of camera and the positional prior of LED are fully explored and utilized;
    \item The performance on simulation and real-world camera-based tactile sensor is comprehensively evaluated.
\end{itemize}

The rest of this paper is organized as follows: Sec. 2 introduces the related works, Sec. 3 describes the theory of the proposed method, Sec. 4 presents the simulated and real-world experiments, and Sec. 5 concludes the contributions, limitations, and future works of this paper.

\section{RELATED WORKS}
\subsection{Camera-based Tactile Sensors}
Since camera-based tactile sensors are proposed~\cite{johnson2009retrographicOriPS}, researchers have improved their hardware and algorithm via various aspects to adapt to robot applications. 
To reduce the sensor size, compact optical designs are introduced, such as waveguide~\cite{li2014localizationGelsight}, declining illumination~\cite{dong2017improvedGelsight}, mirror reflection~\cite{wang2021gelsightwedge}, stereo vision~\cite{zhang2023gelstereo}, and complex lens~\cite{taylor2022gelslim3}. Besides, lookup table based geometry reconstruction methods are proposed~\cite{dong2017improvedGelsight} and improved by neural network~\cite{wang2021gelsightwedge}. Based on geometry measurements, researchers can estimate the force~\cite{yuan2015measurement,zhang2022deltact} and distribution~\cite{ma2019densegelslim2}. 
To further increase the perception range, non-planar designs are proposed, such as round~\cite{gomes2020geltip}, fingertip~\cite{lepora2022digitac}, and cone shape~\cite{tippur2023gelsight360}.
With the development of embodied AI and humanoid robots, researchers try to develop camera-based tactile sensors for dexterous hand, such as soft endoskeleton and palm perception~\cite{liu2024passivelyromeo}, svelte proprioceptive sensing~\cite{zhao2023gelsightsvelte}, fully-covered hand~\cite{romero2024eyesight}, as well as for soft robot and compliant applications~\cite{ren2024softac} and five-finger dexterous hand~\cite{zhiyuan2024hand}.
These efforts improve the sensor integration level for robot applications, such as perception~\cite{yuan2018highPhD}, grasping~\cite{wang2021gelsightwedge}, and manipulation~\cite{bauza2024SimPLE,yuan2022servo}.

\subsection{Tactile Geometric Reconstruction}
\textbf{Look-Up Table.} For fast perception, colorful (red, green, blue, even white) LEDs are mounted on different positions to illuminate the gel surface. Then, a ball presses the gel and calibrates the mapping between gradient and the observed color intensity on the deformable surface of sensor~\cite{yuan2017gelsight}. Then, base on the observed intensity of the pressed deformation surface, one can get the gradient of each pixel by explicit look-up table or implicit models~\cite{wang2021gelsightwedge}. 
However, the computing-complexity will increase with the image resolution, the look-up table size or the model scale, and hard to achieve real-time sensing. Furthermore, the sensor-wise calibration lacks of generality, and the accuracy is affected by nonuniform light intensity. For dexterous grasping and precise manipulation, the tactile accuracy is significant. 

\textbf{Model-based.} For accurate perception, the pioneers~\cite{johnson2011microgeometryProtable} take the classical photometric stereo~(PS) methods to reconstruct the normal of the deformation surface. PS methods~\cite{shi2019benchmark} based on more strict theory and have high accuracy, but the sequential observation process~(under tens of lights) is time-consuming.
Robots need more efficient sensing to provide dynamic information. Researchers try to reduce the number of lights~\cite{zheng2019spline} and the complexity of algorithms.
Symmetric-PS~\cite{minami2022symmetric,makabe2023nearsymme} takes symmetric positions prior of LEDs to solve the surface normal with fewer lights and observations, but they still need multiple sequential observations~(at least 4).
RGB-PS\cite{chakrabarti2016singleRGBPS} uses colorful lights, and splits independent channels from single observation to restore the surface normal.
However, RGB-PS is time-consuming~(minute level). and the channel cross-talk is inevitable, especially the aliasing of R-G and B-G.
How to simultaneously achieve high-efficient perception and ensure the reconstruction accuracy is still a big challenge.

\section{METHOD}
To achieve the highly efficient reconstruction for camera-based tactile sensors, based on the theory of photometric stereo we employ the symmetric illumination to minimize the number of lights. After that, we utilize the spectral features of LEDs and the response curve of camera, and design two group orthogonal red-blue illuminations and take two-shot observations. Then, we split the red and blue channels as four independent observations. Finally, we build a closed-form theoretical method to solve the surface normal with low complexity. Based on the albedo characteristics of camera-based tactile sensor, we can further simplify the proposed method as pure additional manipulation. 
\begin{figure}[!t]
    \centering
    \includegraphics[width=\columnwidth]{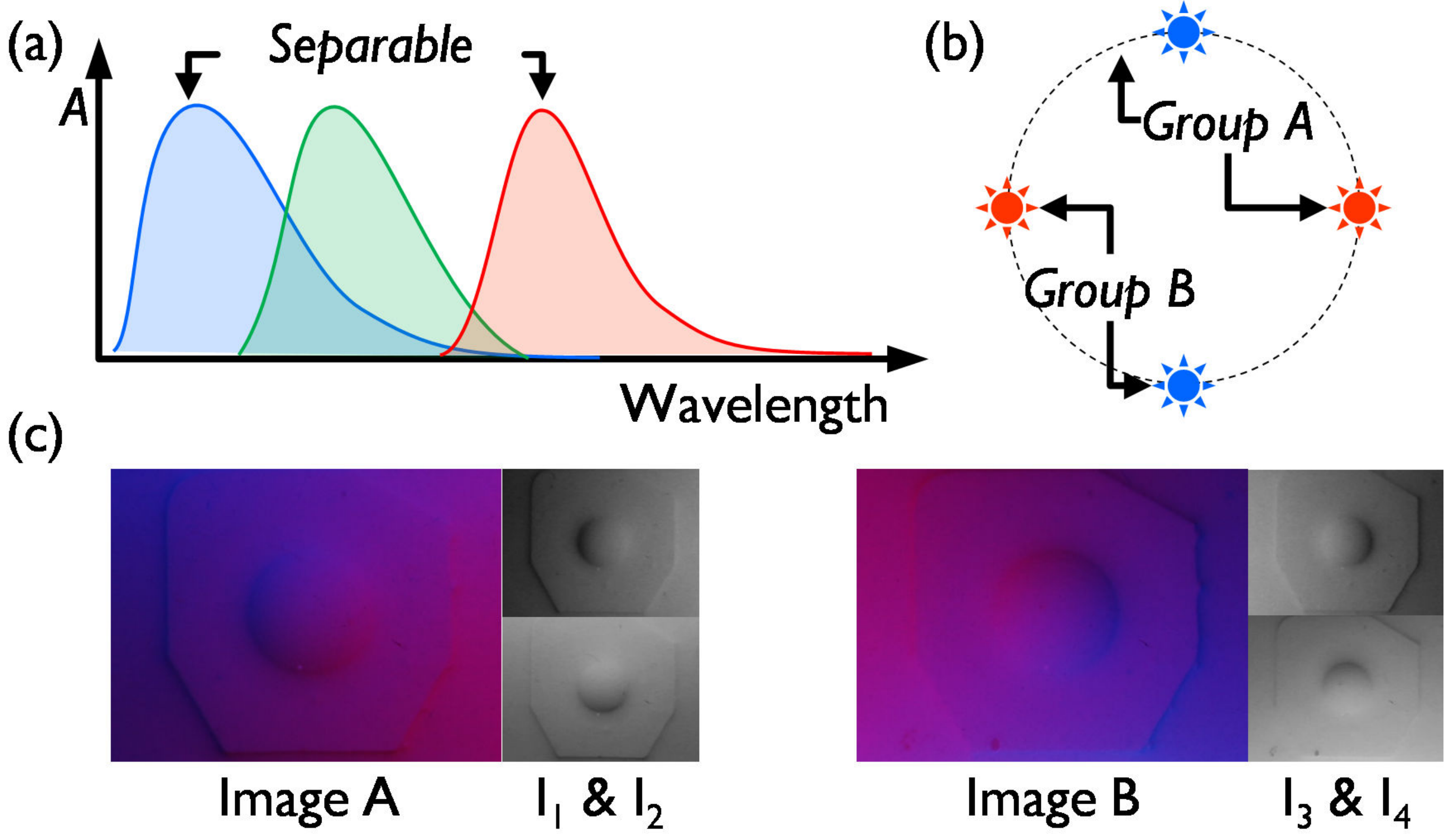}
    \caption{Proposed two-shot photometric stereo with two-shot illuminations. By utilizing the spectral characteristics of chromatic CMOS (red and blue channel has no (little) crosstalk) and LEDs in (a) and their predefined position in (b), red and blue channels can be split from image as (c).\vspace{-0.3cm}}
    \label{fig:spectrum}
\end{figure}

\subsection{Two-Shot Photometric Stereo with Symmetric Prior}\label{sec:two-shot}
Photometric Stereo can model the reflection characteristics of target surface by observing images $[I_0, I_1, ..., I_n] \in \mathbf{I}$ under different illuminations, and can reconstruct the surface normal map from observations.
The imaging process of the pixel $i$ can be represented as:
\begin{equation}
    I_{ij} = \rho f(\mathbf{n}_i,\mathbf{l}_j,\mathbf{v})\mathbf{n_i}^T \mathbf{l}_i,
    \label{eq:img}
\end{equation}
where $\rho$ describes the surface albedo, $f$ is the reflection model, $I_{ij}$ is the observed intensity of $i_{th}$ pixel under $j_{th}$ illumination, the $\mathbf{n}_i \in \mathbb{R}^3$ is the surface normal vector of the point.
$\mathbf{l}_j \in \mathbb{R}^3$ is the $j_{th}$ light. $\mathbf{v}$ is the observation vector ([0, 0, 1], generally). PS need to reconstruct the surface normal based on observations and the light information:
\begin{equation}
    n_i = \mathcal{F}^{-1}(\mathbf{I_i}, \mathbf{L}),
\end{equation}
where $\mathbf{I_i}$ is the observations under light set $\mathbf{L}$.

Based on the characteristics of camera-based tactile sensors, we can assume directional illumination, an orthographic camera, and Lambertian reflection during the imaging process to simplify the analysis. We can rewrite \eref{eq:img} as $I_i = \rho \mathbf{l}_j^T \mathbf{n_i}$.
Classical PS method~\cite{shi2019benchmark} can accurately solve surface normal map with enough lights and observations. 

However, sequential observations under many lights is unpractical in robot tactile sensor.
It is necessary to reduce the lights and increase efficiency. Symmetric lights and chromatic CMOS provide a potential solution.

For symmetric prior, we assume the lights are symmetrically placed on a same height: $[\pm l_x,\quad \pm l_y,\quad l_z]$, which can simplify the imaging representation.
For CMOS spectrum, the Red and Blue channels have little overlap, as shown in Fig. \ref{fig:spectrum}. We can split four observations with two-shot illumination. This further improve the observation efficiency.
Based on this idea, we design a symmetric illumination with independent-spectrum LEDs~(red, blue) as \Tref{tab:light2}.
\begin{table}[!t]
    \centering
    \caption{Symmetric illumination setup in the two-shots case.}
    \begin{tabular}{c|ccc}
    \toprule
        Index & Color & Direction & Direction~(unit distance) \\
        \midrule
        1$^\#$ & Red & $[l_x, 0, z_0]^\top$ & $[1, 0, z_0]^\top$ \\
        2$^\#$ & Blue & $[0, l_y, z_0]^\top$ & $[0, 1, z_0]^\top$  \\
        3$^\#$ & Red & $[-l_x, 0, z_0]^\top$ & $[-1, 0, z_0]^\top$  \\
        4$^\#$ & Blue & $[0, -l_y, z_0]^\top$ & $[0, -1, z_0]^\top$  \\
        \bottomrule
    \end{tabular}
    \label{tab:light2}
\end{table}

Specifically, we simply place the lights on the same symmetric distance~(around the camera) on the height $z_0$, and divide them as two orthographic group. Each group illuminates the deformation surface alternately. \\
\textbf{\textsc{Group 1}}:  {\color{red}$\leftarrow [1,0,z_0]$} and {\color{blue}$\downarrow [0,1,z_0]$}, lighting up for first image capturing (arrow means light direction);\\
\textbf{\textsc{Group 2}}:  {\color{red}$\rightarrow [-1,0,z_0]$} and {\color{blue}$\uparrow [0,-1,z_0]$}, lighting up for second image capturing;

Assuming Lambertian reflectance and no crosstalk between color channels, the observations for light group 1 can be split from the read and the blue channel from image
\begin{equation}
\begin{aligned}
    I_1 &= \rho_r (n_x + N_z), \quad\quad
    I_2 &= \rho_b (n_y + N_z),
\end{aligned}
\end{equation}
where $\boldsymbol{n} = [n_x, n_y, n_z]^\top$, $N_z = z_0 n_z$, $\rho_r,\rho_b$ are the albedo of red and blue channel. The image observations when LED3 and LED4 are turned on are
\begin{equation}
\begin{aligned}
    I_3 &= \rho_r (-n_x + N_z),  \quad\quad
    I_4 &= \rho_b (-n_y + N_z). \\
\end{aligned}
\end{equation}

Based on the above equations, we have
\begin{equation}
\begin{aligned}
    \rho_r n_x &= \frac{I_1 - I_3}{2},  \quad\quad
    \rho_b n_y &= \frac{I_2 - I_4}{2},\\
    \rho_r N_z &= \frac{I_1 + I_3}{2},  \quad\quad
    \rho_b N_z &= \frac{I_2 + I_4}{2}.
\end{aligned}
\end{equation}
Based on the expression on $N_z$, we can derive
\begin{equation}
\begin{aligned}
    \kappa = \frac{\rho_b}{\rho_r} &= \frac{I_2 + I_4}{I_1 + I_3},\\
    \rho_b n_x = \kappa \rho_r n_x &= \frac{(I_2 + I_4)(I_1 - I_3)}{2(I_1 + I_3)}.
\end{aligned}
\end{equation}

We can calculate the surface normal based on photometric stereo, with two-shot, symmetric and colorful observations. 

\paragraph{Case A}: channel-wise albedo.
If $\rho_r \neq \rho_b$, we get an albedo-scaled surface normal represented as:
\begin{eqnarray}
\boldsymbol{b} = 
    \rho_b \begin{bmatrix}
    n_x \\
    n_y\\
    N_z
    \end{bmatrix} = 
    \left[
    \frac{(I_2 + I_4)(I_1 - I_3)}{2(I_1 + I_3)}, 
    \frac{I_2 - I_4}{2},
    \frac{I_2 + I_4}{2}\right]^\top.
\end{eqnarray}
By normalizing $\boldsymbol{b}$ and eliminate the height $z_0$, we get the estimated surface normal (as ST-A).

\paragraph{Case B}: each channel has same albedo.
If $\rho_r = \rho_b = \rho_0$, then $\kappa=1$, we can rewrite the equation as the function of direct observations:
\begin{eqnarray}
\boldsymbol{b} = 
    \rho_0 \begin{bmatrix}
    n_x \\
    n_y\\
    N_z
    \end{bmatrix} = 
    \left[
    \frac{I_1 - I_3}{2}, 
    \frac{I_2 - I_4}{2},
    \frac{I_1 + I_3}{2}\right]^\top.
\end{eqnarray}
The $\rho_0$ is the albedo constant of the target urface (as ST-B).

It is worth to noting that, for this channel-consistent albedo, we only \textbf{require 3 pure additional operations}~(subtraction is special addition) can calculate the normal vector.

\begin{figure}[!t]
    \centering
    \includegraphics[width=0.9\columnwidth]{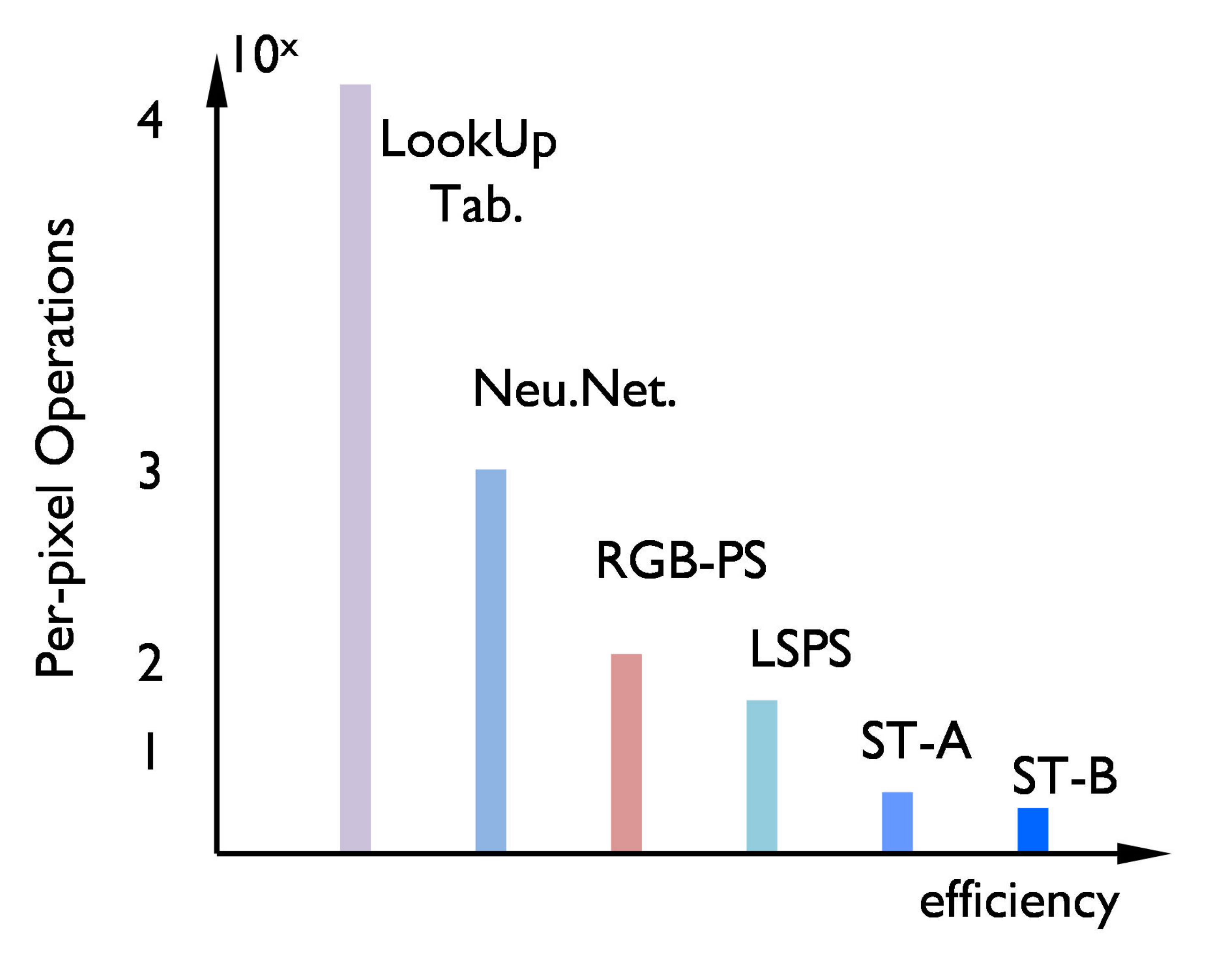}
    \caption{The comparison of computational complexity between typical algorithms for camera-based tactile sensor and our methods (ST-A, ST-B).\vspace{-0.3cm}}
    \label{fig:Ops}
\end{figure}
\subsection{Analysis of Computational Efficiency}
Proposed method not only can reduce the observations time, but also greatly reduce the model complexity. Reducing the capturing time and the simplifying the model computing overhead can collaboratively improve the reconstruction accuracy of camera-based tactile sensor.

\begin{figure*}[!ht]
    \centering
    \includegraphics[width=\textwidth]{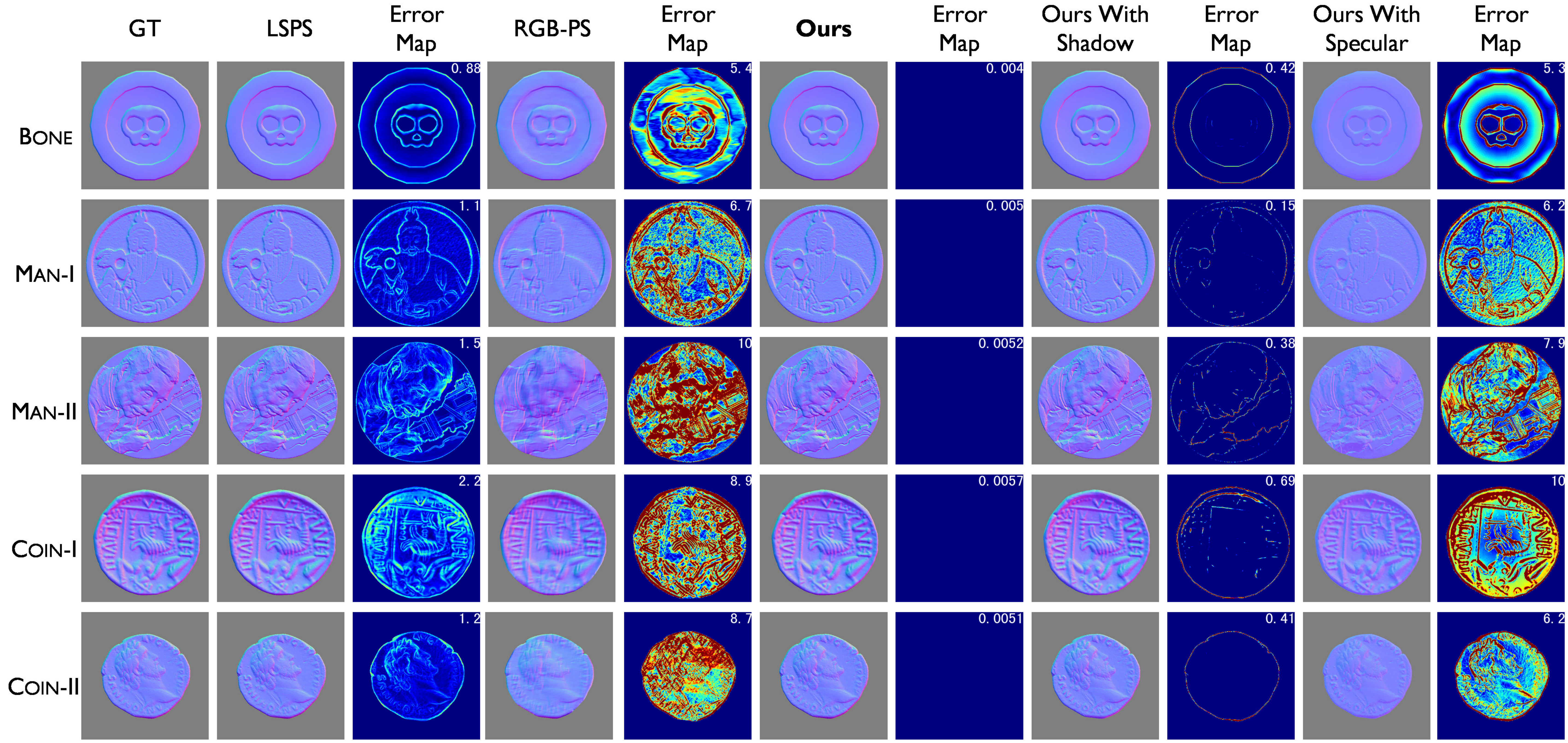}
    \caption{Reconstruction results from typical photometric stereo algorithms and the proposed method. We choose five relief objects to represent the challenging tactile surface (1-7 \textit{cols}). We also introduce spatial-varying and channel-wise albedo in the rendering process. The robustness under shadows and specular surface is evaluated (8-11 \textit{cols}). The mean angular error is attached on the left corner of each error map~(color range is [0$^\circ$, 10$^\circ$]). \vspace{-0.3cm}}
    \label{fig:simComp}
\end{figure*}

To clearly present the calculation efficiency of SymmeTac~(ST-A/B for two case), we compare the computational complexity per pixel. Here, we select representative methods: LSPS~\cite{woodham1980photometricLSPS} and RGB-PS~\cite{chakrabarti2016singleRGBPS} from photometric stereo, Look-up table~\cite{taylor2022gelslim3} and Neural Networks~(NN)~\cite{wang2021gelsightwedge} methods from camera-based tactile sensor. The comparisons are presented in Fig. \ref{fig:Ops}.

For geometry reconstruction, the gradient and the surface normal are functionally equivalent (easily convert to each other by formulas). Complex learning-based methods are not involved, because their large computational overhead is not suit for the robot application on limited hardware. 

Specifically, LSPS takes $n$ observations under $n$ lights, then uses least-square\cite{woodham1980photometricLSPS} to calculate the surface normal vector. Only considering the multiple operation, it needs $12m + 36$ operations~(where $m = 4$ for four lights).
RGB-PS takes patch-wise least-square method to generate local normal vectors, which then are harmonized to produce a global consistent surface normal map. Only considering the local part, it also involve $12m + 36$ operations~(where $m = 3$ for RGB lights).
Look-up table method need a pre-defined calibrated table to mapping the RGB value to gradient (Gx, Gy), the calibration process is laborious and may introduce artificial errors. Even without taking into account the calibration and table building, the computational complexity lookup table operation are $\mathcal{O}(p)$ ($p$ is the elements in the table, such as 90x90x90). Improving accuracy by incorporating more elements leads to longer processing times and larger storage.
Neural network (NN) can replace the table. However, NN needs dense matrix process, not mentioned the pre-process and memory swapping. Typical NN\cite{wang2021gelsightwedge} needs 2272 multiplications and 2242 additions in the matrix calculation.
For our SymmeTac, the two-shot photometric stereo method takes full use of the illumination prior and the low crosstalk of CMOS channels, and greatly decreases the computational overhand. For the spectrum-variable albedo, we need 4 additional and 2 multiple/divide operation (ST-A). For the spectrum-constant albedo, we only need 3 additional operations.

\section{EXPERIMENT}
We design simulated and real-world experiments to evaluate the proposed method. We first test and compare the accuracy and robustness of proposed two-shot photometric stereo method with typical PS method. 
We then implement a real-world symmetrically illuminated camera-based tactile sensor and integrate it with a robot. Extensive experiments are conducted to demonstrate the effectiveness and potential of this novel technique.
Because the surface normal map describes the detailed geometry of the target, allows for comparison with other PS methods, and is easy to convert into gradient and depth maps (as classical approaches), we will focus on its analysis in the following sections.

\subsection{Simulation Experiments}
We first simulate the illumination and render images under different surface reflectance or global illumination effects. Then, we choose typical PS algorithms to evaluate their accuracy on different situations, as shown in Fig.\ref{fig:simComp}.

We collect 5 objects~(reliefs is related to the tactile target) with diverse shapes. To theoretically demonstrate the effectiveness of our method, we first render a dataset named {\sc IdealSymData} under Lambertian reflectance and ignore the shadows and global illumination effects such as inter-reflections, where the surfaces are illuminated by $4$ symmetric normalized light directions:\\
\hbox{$[\frac{1}{\sqrt{2}}, 0, \frac{1}{\sqrt{2}}$]}, \, [$0, \frac{1}{\sqrt{2}}, \frac{1}{\sqrt{2}}$], \; [$-\frac{1}{\sqrt{2}}, 0, \frac{1}{\sqrt{2}}$],\, [$0, -\frac{1}{\sqrt{2}}, \frac{1}{\sqrt{2}}]$.

\noindent In addition, to test our method under specularities and shadows, we use the Cycles engine in Blender to render {\sc BlenderSymData}, leading to $75$ samples for evaluation. 

\paragraph{Baseline \& Metric.} As our method is based on photometric stereo settings, we choose the classical least-square based photometric stereo~\cite{woodham1980photometricLSPS} as our baseline, abbreviated as ``LSPS''. In addition, to compare photometric stereo with color light settings, we choose RGB-PS~\cite{chakrabarti2016singleRGBPS}. We choose the image observations under top, right, and left light to compose an RGB image and then take it as the input for RGB-PS. Following the evaluation manner from existing photometric stereo methods, we use mean angular error~(MAE) in degree as the metric for measuring the difference between the GT and estimated surface normals.

\paragraph{Evaluation on Normal Estimation.}
As shown in \fref{fig:simComp}, we compare LSPS~\cite{woodham1980photometricLSPS}, RGB-PS~\cite{chakrabarti2016singleRGBPS} with our method on {\sc IdealSymData}. Since the surface is covered by spatially-varying albedo, which violates the albedo assumption of RGB-PS~\cite{chakrabarti2016singleRGBPS}~(the spatial distribution of albedo can be grouped into limited clusters.) Also, LSPS~\cite{woodham1980photometricLSPS} assumes the albedo to be the same at different color channels, which is invalid in our color illumination setting. Therefore, our method achieves the smallest mean angular error on normal estimation compared with the baselines (col $6^{th},7^{th}$). In addition, the MAE is smaller than 0.001 degree, demonstrating the theoretical effectiveness of our method, where the residual error is mainly caused by image quantization noise.

\paragraph{Robustness against Shadows and Specular Highlights}
As shown in \fref{fig:simComp}, we test the robustness of our methods against shadows and specular highlights, which are not explicitly handled in our method but commonly exist in real-world scenarios. Compared with the error distribution shown in \fref{fig:simComp}, col $8^{th},9^{th}$, the result from our method under cast and attached shadows becomes less accurate in regions where shadows occur. On the other hand, when the surface contains specular highlights, the normal estimation error increase since we assume Lambertian reflectance. On the near-planner surface shown in \fref{fig:simComp}, col $10^{th},11^{th}$, the proposed method is more sensitive to the specular highlights compared to the shadows. 

\begin{figure}[!t]
    \centering
    \includegraphics[width=0.8\columnwidth]{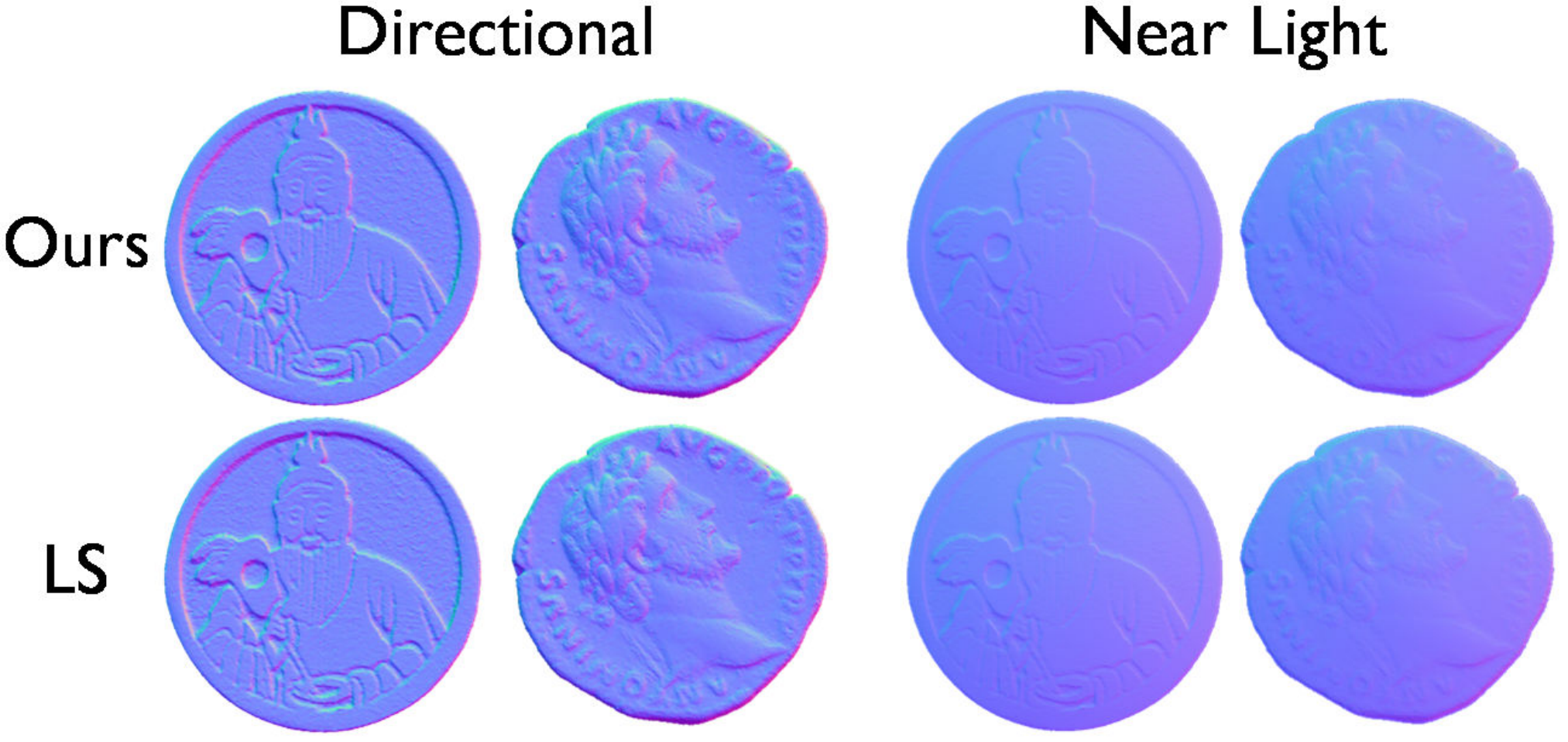}
    \caption{Results from our two-shot PS and classical LSPS method under both directional illumination and near-light illumination. The normal map show slight bulge caused by near-light nonuniform and non-parallel illumination.\vspace{-0.3cm}}
    \label{fig:nearLight}
\end{figure}
\paragraph{Distortion caused by Near-light illumination.} 
In camera-based tactile sensor, the small sensor size may lead to different incident angle and nonuniform illumination.
To further investigate this near-light illumination in simulation, we put the object in origin and normalize its diameter as 1 m. Then we put the lights at $[1, 0, 1], [0, 1, 1], [-1, 0, 1], [0, -1, 1]$ and place the camera at $[0,0,2]$ with unit m.
As shown in Fig. \ref{fig:nearLight}, due to the spatial variability of light intensity and incident direction, pixels closer to the light source in the observed area receive stronger illumination and smaller incident angles. Consequently, the top-left region tends to appear blue, while the bottom-right region shifts towards red in the reconstructed results (more towards the x direction). We also compare the LSPS method, demonstrating that the performance of both algorithms degrades obviously under near-field illumination. These results provide a solid explanation for surface normal map distortions in real-world sensor.

\subsection{Sensor Design}  
\label{sec:sensor}
\begin{figure}[!t]
    \centering
    \includegraphics[width=\columnwidth]{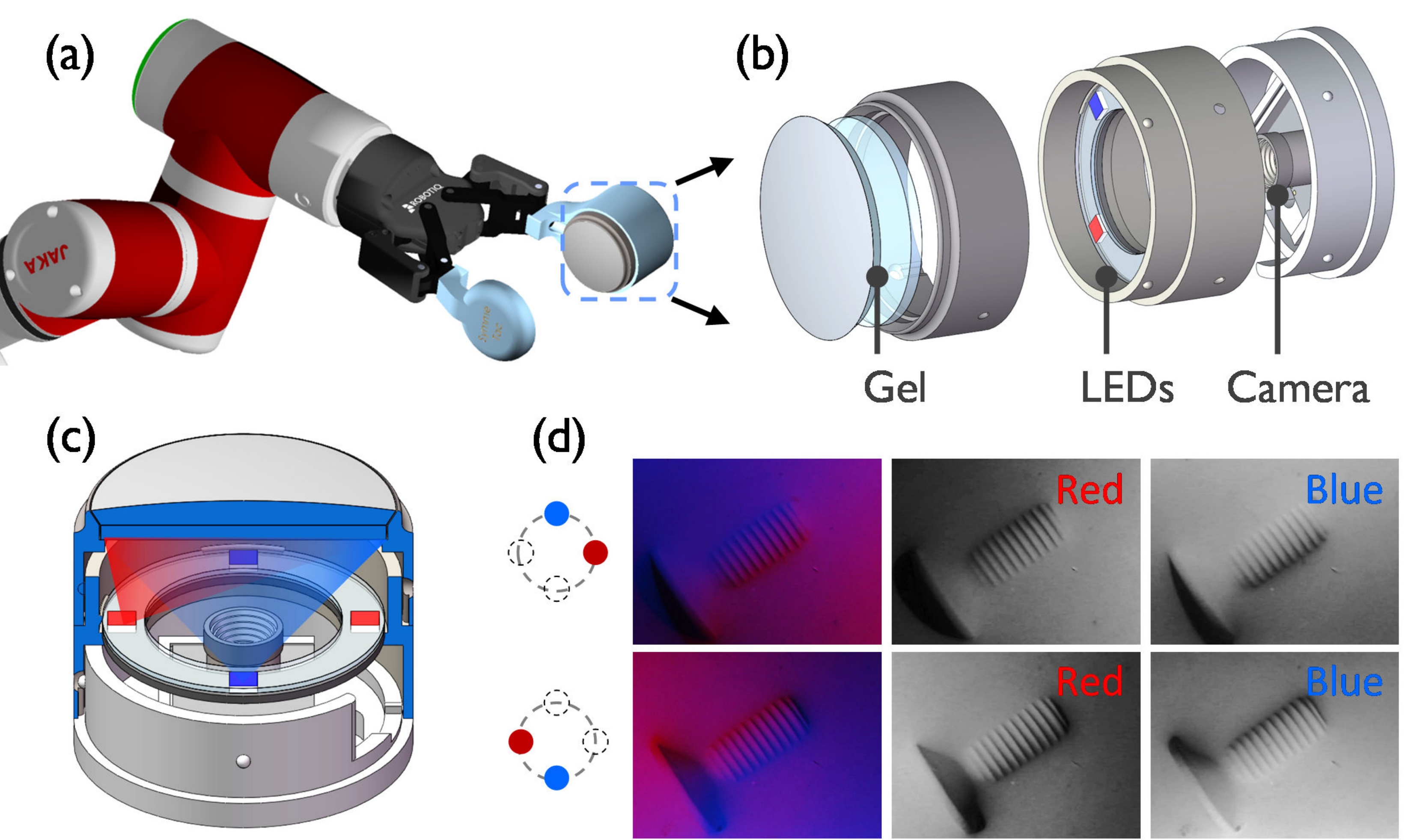}
    \caption{The structure of camera-based tactile sensor with symmetric illumination. (a) The sensor is integrated on the Robotiq gripper, then mounted on the Jaka robot arm; (b) Proposed tactile sensor design with four symmetric LEDs; (c) The section view and (d) the observed two-shot images and their red/blue channel.\vspace{-0.4cm}}
    \label{fig:sensorStructure}
\end{figure}

With the previous modeling and analysis, we can design a novel camera-based tactile sensor with symmetric LED setup and color configuration.
We choose the cylinder structure as sensor shape for symmetric illumination, and design a sandwich structure for the high-efficient sensor. There are three main structures to host the camera, LEDs and the soft gel surface for perception. All structure are fabricated by 3D print with PLA, as well as the mold of Gel layer (transparent 00-30 silicon gel).
Proposed sensor is mounted on the Robotiq 2F-85 right gripper, and an assistive palm is mounted on the left gripper.
These modules are mounted on the JAKA Zu 7 robot arm. 
Details are shown in \Fref{fig:sensorStructure}.

\begin{figure}[!t]
    \centering
    \includegraphics[width=0.8\columnwidth]{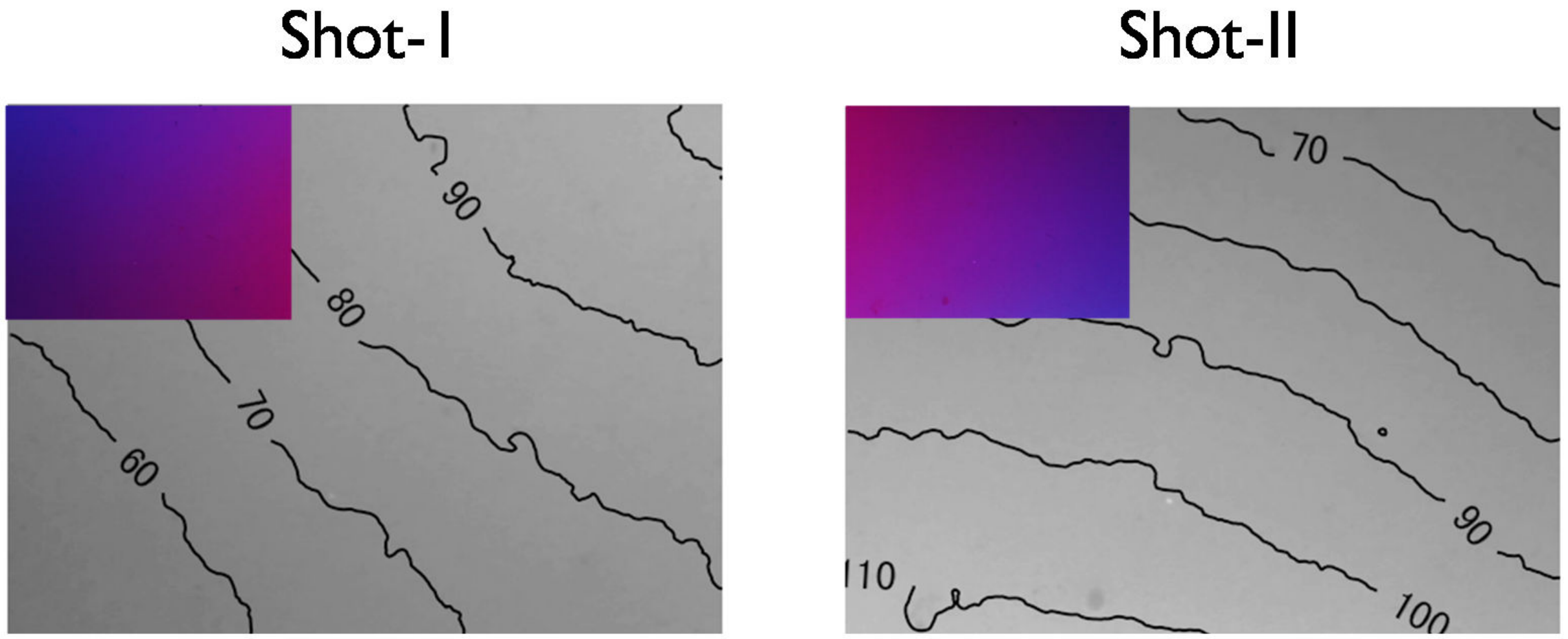}
    \caption{Illumination quantitative evaluation. The upper-left part shows the observed image, and the right part shows the intensity distribution under each shot. The proposed sensor design has uniform intensity with small standard deviation (11.9 of shot-1 and 12.7 of shot-2 under [0,255]).\vspace{-0.3cm}}
    \label{fig:evalInts}
\end{figure}
\paragraph{Illumination Design and Robot Integration.}
As theoretical analysis, we split 4 LEDs (on the angle of 0$^\circ$, 90$^\circ$, 180$^\circ$, 360$^\circ$) as two group, each group has red and blue light. 
We choose a USB camera with 640x480 resolution and choose the Arduino UNO to control LEDs.
Especially, 4 programmable WS2812 LEDs ring is mounted in the middle layer, which has 620-625 nm red light and 465-467 nm blue light, can help to avoid the cross-talk, as shown in Fig. \ref{fig:spectrum}. 

The Arduino UNO controller can send the light color and switch command to each light. 
The synchronous capture, illumination and reconstruction program is running in the computer.
We quantitatively evaluate the illumination feature. The intensity contours of each channel are depicted in Fig. \ref{fig:evalInts}. The illumination intensity is uniform in the field-of-view. The $\sigma$ of each shot is smaller than 5\% (11.9 of shot-1 and 12.7 of shot-2 under [0,255] range).

\begin{figure}[!t]
    \centering
    \includegraphics[width=\columnwidth]{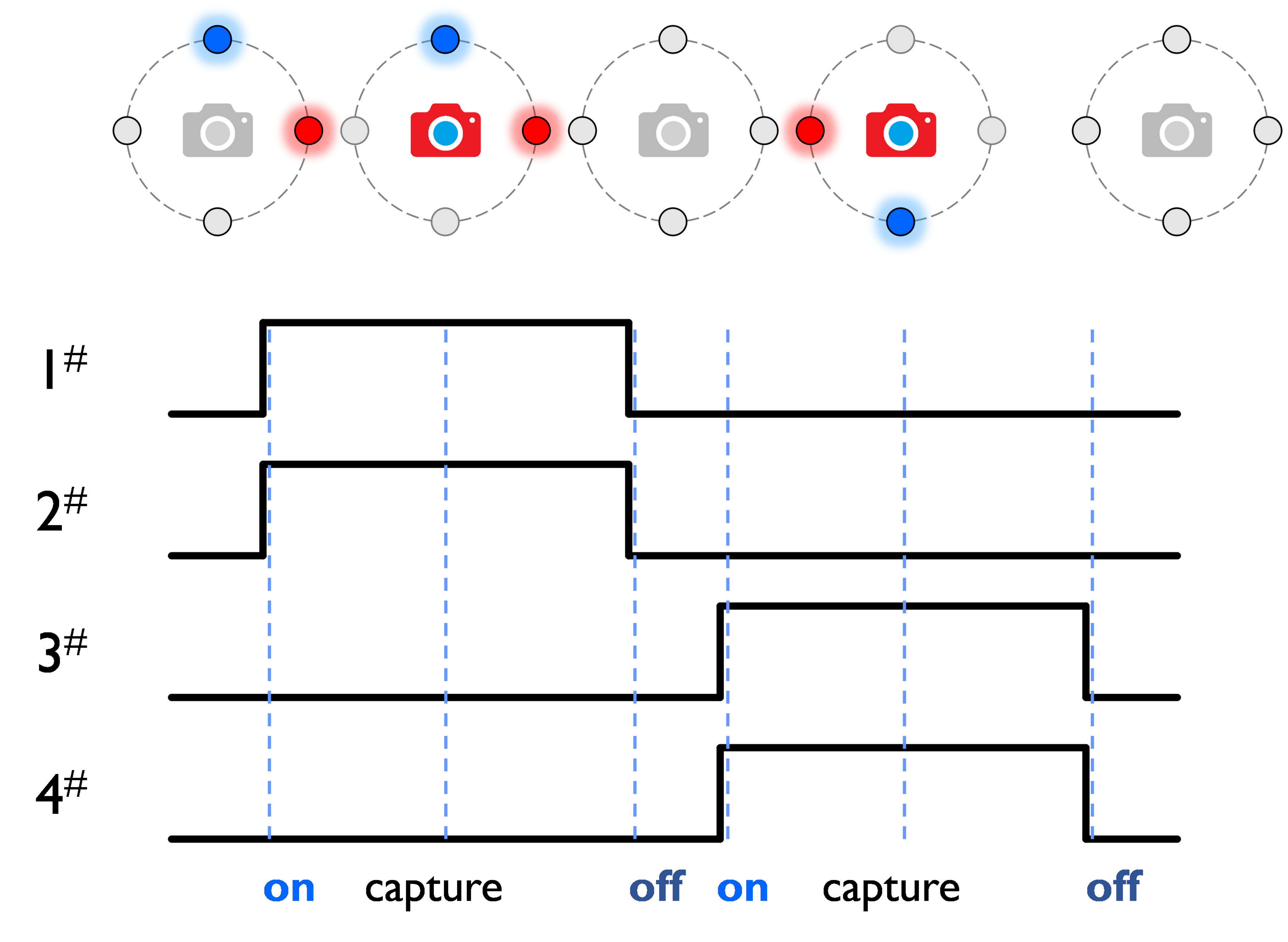}
    \caption{Observation timing diagrams of SymmeTac. Two-shot illuminations and observations are organized as a period. In each observation period, 1$^\#$ red and 2$^\#$ blue first turn on and capture the first image; 3$^\#$ red and 4$^\#$ blue then turn on and capture the second image. Because the lighting and capture is fast, we can achieve highly efficient observation and reconstruction.\vspace{-0.3cm}}
    \label{fig:TimingDiagrams}
\end{figure}

\paragraph{Capture Configuration.}
Based on the two-shot photometric stereo model, the tactile sensor takes two observations sequentially.
First, the program turns on the \textsc{group 1} lights, capture the observation $I_A$ under red (180$^\circ$) and blue (90$^\circ$) illumination, and turns off the \textsc{group 1};
Second, the program turns on \textsc{group 2}, captures the observation $I_B$ under red (0$^\circ$) and blue (270$^\circ$) illumination, splits the corresponding channels, and turns off the \textsc{group 2}. 
The timing diagrams are shown in Fig. \ref{fig:TimingDiagrams}.
Based on the $I_A$ and $I_B$, we can split the channels ($I_1$ and $I_2$ from $I_A$, $I_3$ and $I_4$ from $I_B$) and calculate the surface normal with the two-shot photometric stereo theory in Sec. \ref{sec:two-shot}.

\subsection{Real World Experiments}
\begin{figure}[!ht]
    \centering
    \includegraphics[width=\columnwidth]{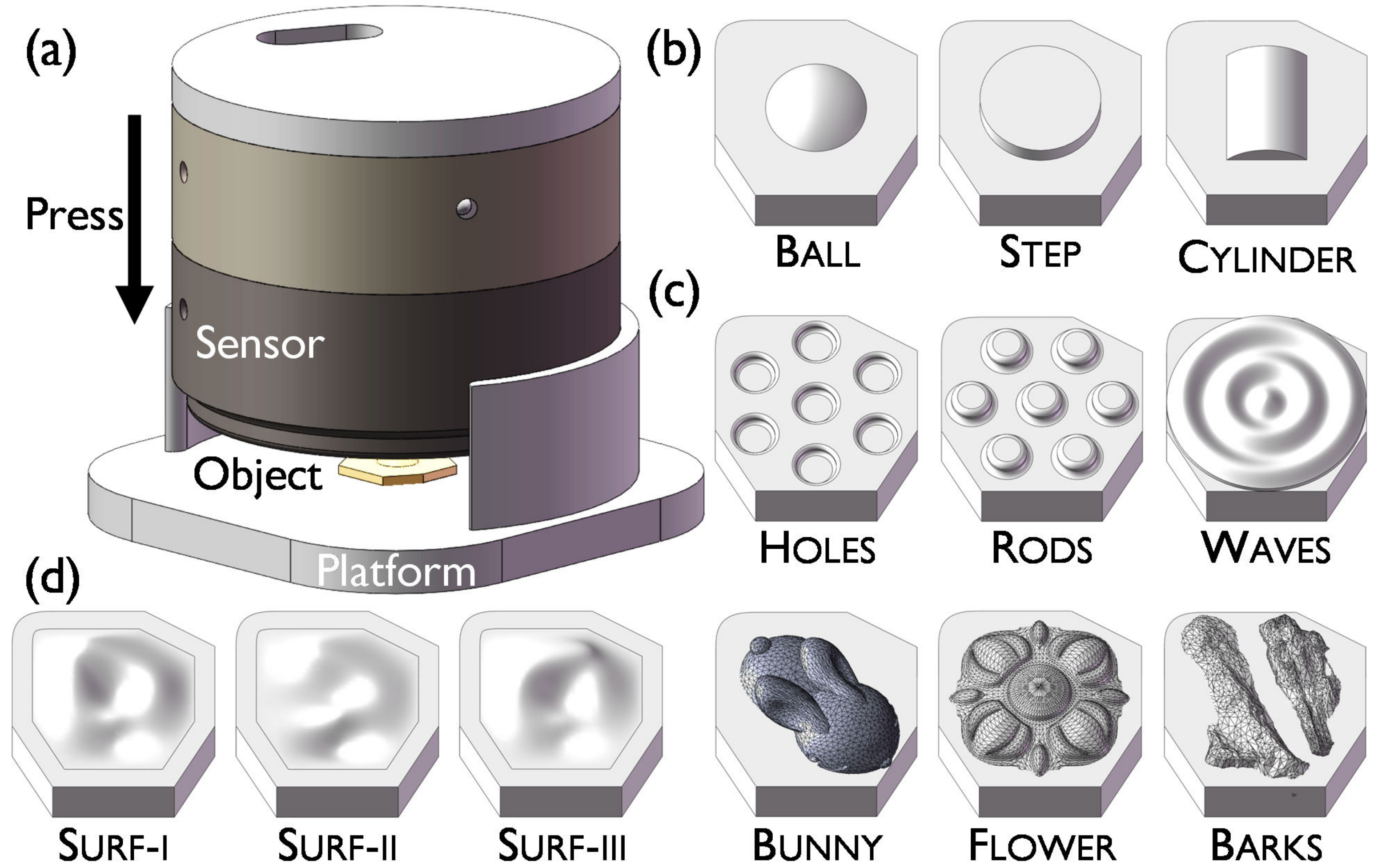}
    \caption{Designed tactile shape mini dataset. Left is the capture platform, where the shape is placed on a locating hole for press and reconstruction testing. Right are the CAD model of design shapes, which include three groups (12 surfaces): \textbf{Simple}~(b), \textbf{Repeat}~(c), \textbf{Complex}~(d).\vspace{-0.3cm}}
    \label{fig:minidatasetCAD}
\end{figure}

To evaluate the performance of proposed sensor, we test the tactile performance on diverse objects with rich surface geometry, and on grasping tasks with dynamic interactions. For the statics perception, we fabricate a mini dataset with 12 typical shapes and complex geometry. For the dynamic interaction, we adopt the gripper to grasp and manipulate typical tools.
Besides, we also collect diverse clothes, fastener and vegetables (fruits) to explore the applications on diverse industrial applications.

\begin{figure*}[!ht]
    \centering
    \includegraphics[width=\textwidth]{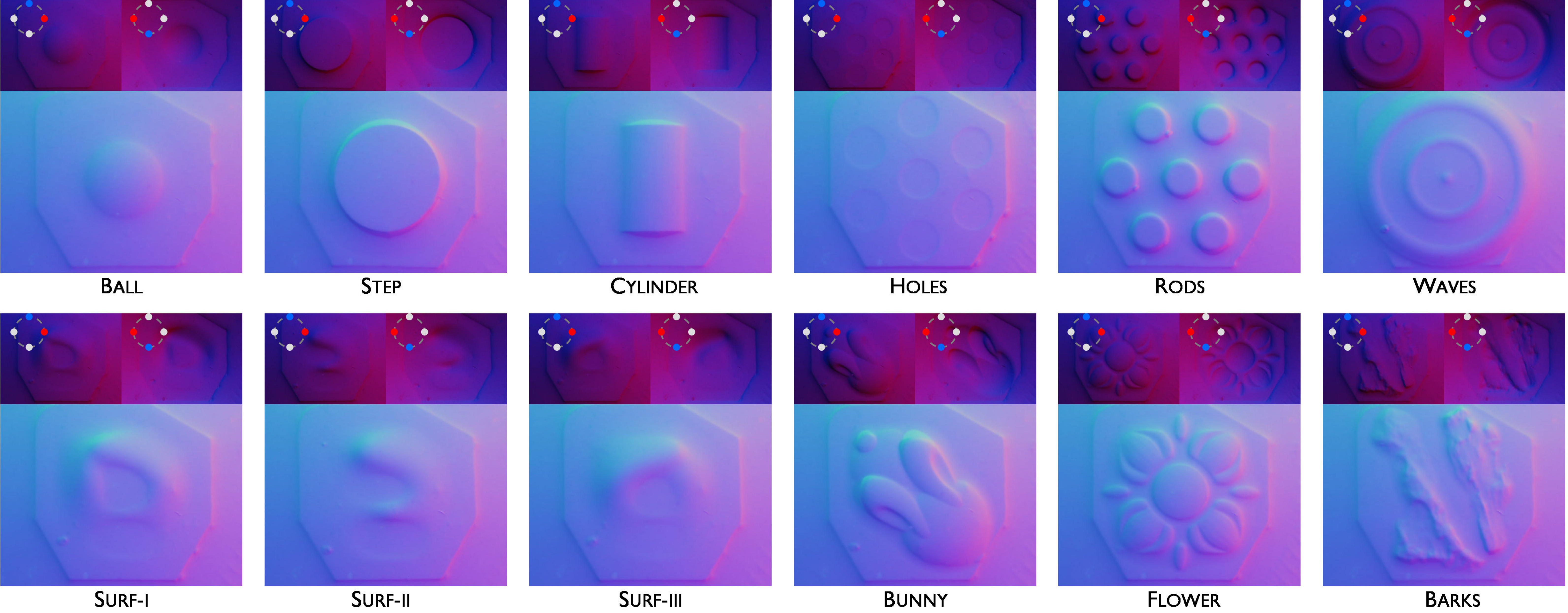}
    \caption{Surface normal reconstruction performance of proposed tactile sensor with two-shot PS. The details from each group objects are clearly restored. Especially, we can clearly observed the surface texture of \textsc{Complex} group.\vspace{-0.3cm}}
    \label{fig:minidatasetEst}
\end{figure*}
\subsubsection{Statics Perception}
The mini geometry dataset contains diverse surface geometries and is divide into three groups, as shown in Fig. \ref{fig:minidatasetCAD}.
\textbf{Simple} includes basic shapes: \textsc{Ball}, \textsc{Step}, and \textsc{Cylinder}. This group can evaluate the basic tactile accuracy of the sensor. 
\textbf{Repeat} includes period geometry and object arrays: \textsc{Holes}, \textsc{Rods}, \textsc{Waves}. The periodical shapes will test the sensor consistence and the lateral resolution. 
\textbf{Complex} contains free-surface and reliefs: \textsc{Surf1-3}, \textsc{Bunny}, \textsc{Flower} and \textsc{Barks}. This group has diverse surface and geometry distribution, can verify the sensor generality and adaptation.

Considering the gel deformation amplitude, the height of these shapes are limited to 1.5 mm.
Each object is placed on a square board with 10 mm edge. One corner is rounded for locating. 
We design the platform to host these work pieces and contact the sensor with predefined direction. We utilize the proposed camera-based tactile sensor to press shapes from the mini dataset for testing.

Two-shot observations and the reconstructed surface normal maps are shown in Fig. \ref{fig:minidatasetEst}.
Results show our sensor can accurately obtain the surface geometry with the help of proposed two-shot PS, which not only exhibit consistence and generality, but also present high robustness on complex structures, which can support diverse geometry requirements of tactile perception.
Because of the size limitation, there is imperfect near-light illumination, which may induce the slight global distortion in the normal map (upper left tends to cyan and lower right tends to pink). These effect will be imporved in the future optical design research.

\begin{figure}[!t]
    \centering
    \includegraphics[width=\columnwidth]{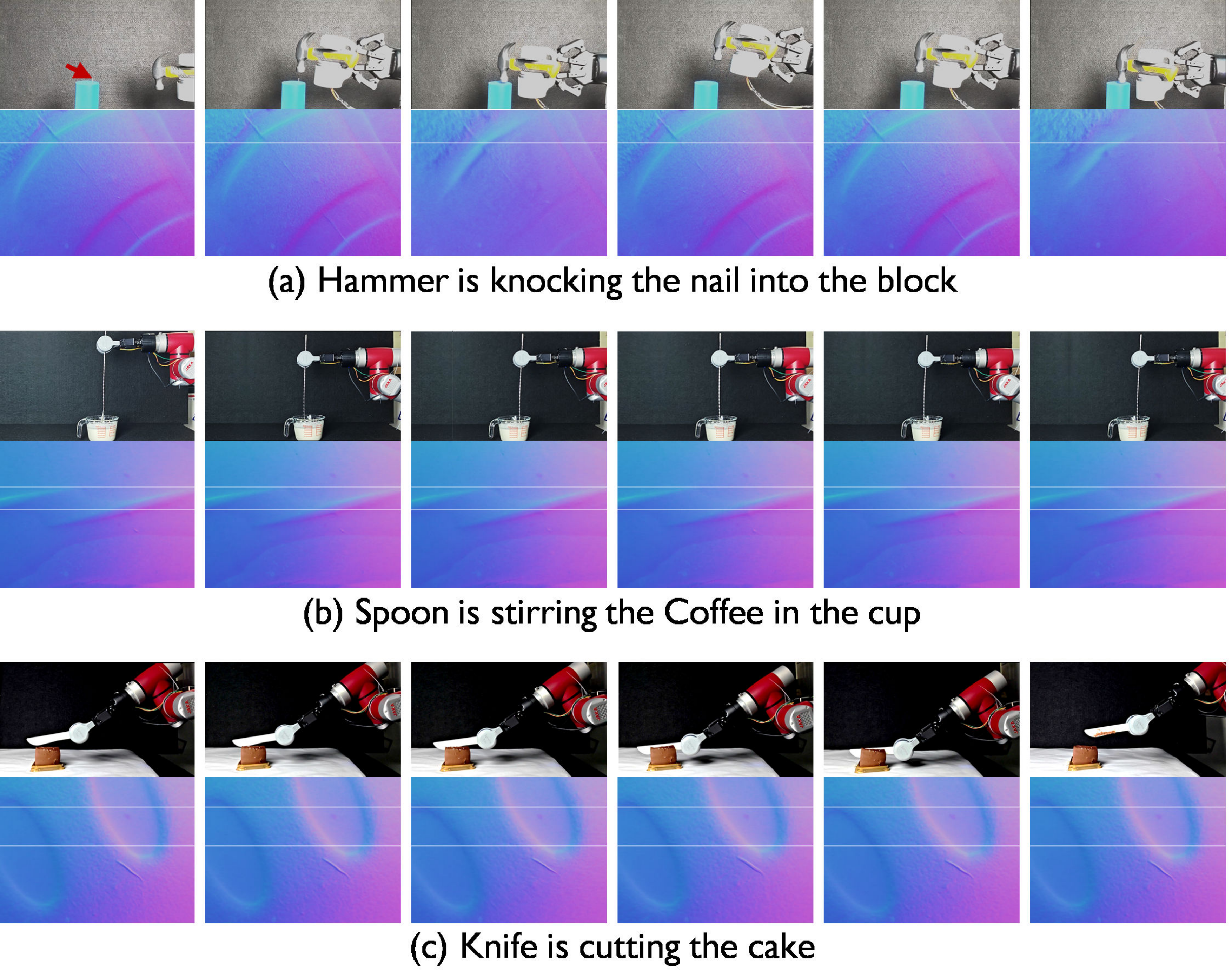}
    \caption{The dynamic tactile results (surface normal map) of tools manipulations. 
    Key frames are presented in each row. White lines assist in identifying the micro movement.
    (a) Knocking the nail with a hammer;
    (b) Stirring batter with a spoon;
    (c) Cutting the cake with a knife.
    More results are available on our \href{https://github.com/Tacxels/SymmeTac}{website}.
    \vspace{-0.3cm}}
    \label{fig:sensortools}
\end{figure}
\subsubsection{Dynamic Tools Grasping and Perception}
High-efficient camera-based tactile sensor can provide dynamic information for robots-object interaction. Based on previously designed gripper in Fig. \ref{fig:sensorStructure}, we present the dynamic tactile perception in robot tools manipulations. 

We choose three typical tools and their corresponding applications in industries and daily life: 
hammer knocks the nail, spoon stirs the liquid, knife cuts the cake. The tactile gripper holds the tool to manipulate the object, we can efficiently restore the subtle dynamic interaction~(geometric changes of contact area), as shown in Fig. \ref{fig:sensortools}. The time-serial results clearly record the change of tactile information, which is significant for further research in controlling and learning for robot dexterous grasping and complex manipulations, especially in the real-world environment.

\begin{figure*}[!ht]
    \centering
    \includegraphics[width=\textwidth]{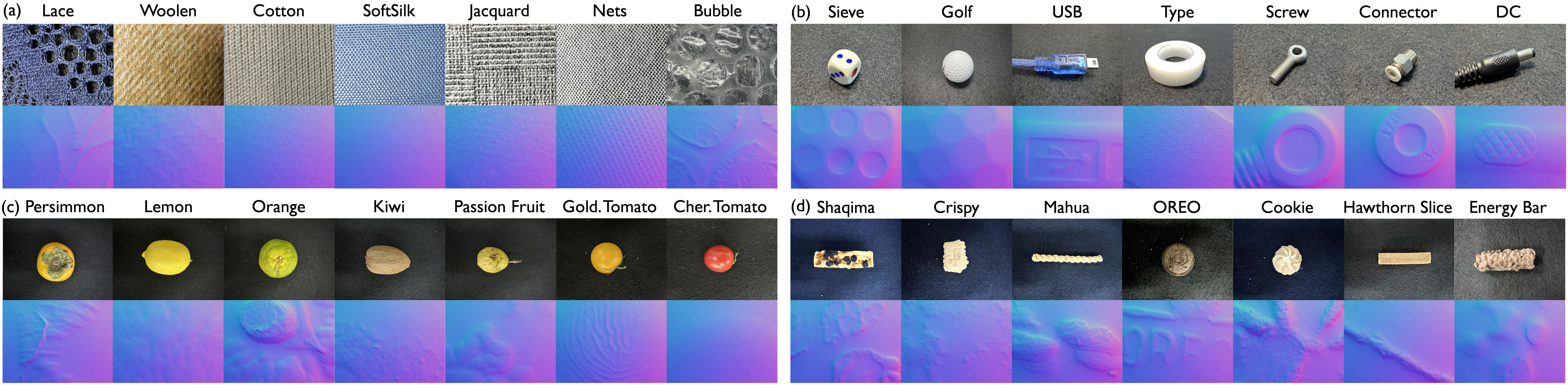}
    \caption{\vspace{-0.2cm}Real-world grasping and surface perception experiments on diverse objects from clothes~(a), industry components~(b), fruits~(c) and snacks~(d). \vspace{-0.0cm}}
    \label{fig:expGene}
\end{figure*}

\subsubsection{General Tactile Applications}
To exhibit the adaptability, we also use diverse daily objects to test its performance. Objects are selected from three ordinary groups: clothes (with rich texture), industry components (such as nuts and bolts), and fruits (with diverse hardness, shape and peels) and crispy food (cookies). The observed images and the reconstructed surface geometry are presented in Fig. \ref{fig:expGene}. 
We can restore the delicate texture of clothes, subtle details on screws and the fragile features of fruits. Especially, the slight surface damage on fruit, the inscriptions on the nuts, and the tiny thread ends of clothes can be captured and reconstructed. We can observe the obvious difference between similar appearance clothes. Besides, distinctive surface structures are captured from cookies. 
These results present the potential applications in diverse industries.

\section{CONCLUSION}
In this paper, we propose a highly efficient photometric stereo method for camera-based tactile sensors, which takes the symmetric prior of LEDs and utilizes the channel separability of CMOS to gracefully solve the surface normal from intensity with simple calculations. Both simulation and real-world sensor experiments show the competitive performance of the proposed method. This method not only can improve the calculation efficiency and the reconstruction accuracy of camera-based tactile sensors, but also provide a novel perspective for photometric stereo theory.

However, the theory still has limitations, which do not fully consider the non-ideal imaging process under point light. 
Besides, the low capture FPS of the low-cost camera and the slightly bulky size of sensor limit the proposed method from fully releasing its potential.

In future research, we will develop a more compact theory for non-ideal light and address global illumination effects for robust tactile sensing. Besides, a high-performance industrial camera will be introduced to achieve fast reconstruction and the mechanical design will be optimized for robot gripper and dexterous manipulation. We believe that this highly efficient and high-performance photometric stereo approach could advance the development of camera-based tactile sensors and robotic perception.







\end{document}